\documentclass[runningheads]{llncs}
\usepackage[T1]{fontenc}
\usepackage{amsmath}
\usepackage{amssymb}
\usepackage{amssymb}
\usepackage{graphicx} % Required for inserting images
\usepackage{algorithm}
\usepackage{algorithmic}
\usepackage{booktabs}
\usepackage{esvect} % Better vector notation, use \vv instead of \vec
\usepackage{subfigure}
\usepackage{float}
\usepackage{hyperref}
\usepackage{cleveref}
\usepackage{comment}
\usepackage{wrapfig}
\usepackage{multirow}
\usepackage{subcaption}
\usepackage{graphicx}
\begin{document}
\title{Markov Chain Decoders Overcome the Heavy-Tail Limitations of Lipschitz Generative Models}
\titlerunning{Markov Chain Decoders for Heavy-Tailed Generation}
\author{Abdelhakim Ziani\inst{1,2}\and
Andras Horvath\inst{2} \and
Paolo Ballarini\inst{1}}
\authorrunning{A. Ziani et al.}

\institute{
Université Paris Saclay, Lab. MICS, CentraleSupélec,  Gif-sur-Yvette, France 
\email{\{hakim.ziani,paolo.ballarini\}@centralesupelec.fr}
\and 
Università di Torino, Torino, Italy\\
\email{horvath@di.unito.it}
}
\maketitle              % typeset the header of the contribution

\begin{abstract}
Heavy-tailed distributions are prevalent in performance evaluation, network traffic, and risk modeling. This behavior poses a fundamental challenge for modern deep generative models. Standard Variational Autoencoders (VAEs) employ Gaussian decoder likelihoods and Lipschitz-constrained neural networks, a combination that is structurally incapable of producing heavy-tailed outputs: the Gaussian tail decays exponentially, and Lipschitz continuity prevents the decoder from amplifying rare events from the latent space input to sufficiently overcome this decay. We provide both a theoretical characterization of this limitation and a controlled empirical demonstration using synthetic Pareto data across a grid of tail indices $\alpha \in \{2, 3, 5, 30\}$ and dimensions $d \in \{1, 5, 10\}$. As a solution, we replace the Gaussian decoder with a Phase-Type (PH) distribution based on Markov chains, while keeping the encoder, latent space, and training procedure identical. PH distributions allow for arbitrarily precise approximations of any positive-valued distributions, including heavy-tailed families. 

Experiments showed that the PH-based model reduces tail Kolmogorov-Smirnov distance by up to $\times6$ and extreme quantile error by up to $\times$10 compared to the Gaussian baseline for heavy-tailed data. These results demonstrate that integrating Markov chain-based distributions into the decoder of a generative model provides a principled, practically effective solution to the heavy-tail generation problem.

\keywords{ Phase-Type distributions \and Variational Autoencoders \and Heavy-tailed distributions \and  Generative models \and Pareto distribution \and Markov chains}
\end{abstract}

\section{Introduction}
\label{sec:intro}
Heavy-tailed distributions are ubiquitous in the systems studied by the performance evaluation community. Network traffic inter-arrival times, file transfer sizes, insurance claim sizes, and financial returns all exhibit slow polynomial tail decay \cite{VogelPapalexiou2025,WuPeng2025,JiaX2014}, meaning that rare but extreme events are far more probable than Gaussian intuition would suggest. Accurately modeling these extremes becomes challenging, in queuing systems, for instance, the heavy-tailed nature of service times fundamentally changes queue length distributions and mean waiting times \cite{AlloucheGirard2024,FossKorshunov2011}.

Deep generative models have emerged as powerful tools for learning complex data distributions. Among these, the Variational Autoencoder (VAE)~\cite{Kingma2013AutoEncodingVB} is particularly attractive due to its probabilistic formulation and tractable training objective. However, standard VAEs exhibit structural limitations when modeling heavy-tailed data\cite{floto2023the}. A key reason lies in the Lipschitz nature of neural networks: encoder and decoder mappings are compositions of bounded-weight layers and Lipschitz activations (e.g., ReLU or tanh), resulting in an overall Lipschitz transformation from the latent space to the data space. When the latent variable follows a light-tailed distribution such as a Gaussian, this Lipschitz mapping cannot generate heavy-tailed outputs with polynomial decay. In practice, the decoder produces a mean function $\mu(z)$ that serves as the reconstruction, and the distribution of generated samples is therefore largely governed by this neural mapping. As a result, the induced data distribution inherits the light-tailed behavior of the latent space rather than exhibiting genuine heavy-tailed structure. The consequence is a form of tail collapse: even after extensive training, a Gaussian VAE cannot place sufficient probability mass in the extreme regions of a heavy-tailed distribution. The reconstruction loss penalizes the model for the bulk of the data, and the Gaussian assumptions prevent the tail from being recovered regardless of how expressive the latent representation becomes.

\textbf{Contributions:}
In this paper, we propose to replace the decoder likelihood with a Phase-Type (PH) distribution\cite{neuts1975}, i.e., the absorption time of a finite-state continuous-time Markov chain, while keeping the architecture, latent space, and training procedure identical. PH distributions, whose approximation properties for heavy-tailed distributions have been well established in the applied probability literature~\cite{Buchholz2014,HorváthTelek2024}, are not subject to the same structural limitations as Gaussian decoders: their shape is determined by the transition structure of the underlying Markov chain, which can be learned end-to-end to match the data, including its tail.

\begin{enumerate}
  \item A theoretical characterization of why Lipschitz generative models with Gaussian decoders cannot produce heavy-tailed outputs.
  \item A controlled empirical study on synthetic Pareto data across tail indices $\alpha \in \{2, 3, 5, 30\}$ and dimensionalities $d \in \{1, 5, 10\}$, isolating the effect of the decoder likelihood from all other architectural choices.
  \item Practical guidance on when Gaussian observation models become unreliable for extreme-value modeling in generative frameworks.
  \item Quantitative evidence that Phase-Type decoders substantially reduce tail distortion for heavy-tailed data.
  
\end{enumerate}

The remainder of the paper is organized as follows. 
Section~\ref{sec:related} reviews related work, 
Section~\ref{sec:background} reviews heavy-tailed distributions, VAEs, and Phase-Type distributions. Section~\ref{sec:theory} provides the theoretical argument for tail collapse. Section~\ref{sec:setup} describes the experimental setup. Section~\ref{sec:results} presents results. Section~\ref{sec:discussion} discusses implications, and Section~\ref{sec:conclusion} concludes.

\section{Related Work}
\label{sec:related}
\subsection{Heavy-Tailed Distributions and Generative Models}

A foundational theoretical result, established across the Generative Adversarial Networks (GAN) and normalizing flow literature, shows that a neural network with Lipschitz-continuous activations fed by a light-tailed input cannot itself produce heavy-tailed outputs~\cite{huster2021pareto,jaini2020tails}. This structural barrier has been approached in several ways. Huster et al.~\cite{huster2021pareto} introduce Pareto GAN, injecting Pareto-distributed noise into the GAN latent space and proving that ReLU networks can preserve regular variation with a controllable tail index. Allouche et al.~\cite{allouche2022ev} propose EV-GAN, reparameterizing the generator via extreme-value theory to approximate heavy quantile functions that standard networks cannot represent. Bhatia et al.~\cite{bhatia2021exgan} augment training data with Generalized Pareto Distribution (GPD)-fitted extreme samples in ExGAN. All these GAN-based approaches require prior knowledge of the tail family or explicit tail index estimation as a preprocessing step.

Within the VAE family, the incompatibility is equally well-documented. Lafon et al.~\cite{lafon2023vae} prove that a standard VAE with a Gaussian prior and decoder produces only light-tailed marginals, and propose ExtVAE, which decomposes data via a multivariate regular variation framework. Kim et al.~\cite{KimKwon2023,Bouayed2025Conditionalt3VAEEL} replace both prior and decoder with Student-$t$ distributions, deriving a modified Evidence Lower Bound (ELBO) based on $\gamma$-power divergence, their tail index must be fixed in advance rather than learned. Floto et al.~\cite{floto2023tilted} address out-of-distribution behavior with a tilted VAE that reweighs the ELBO, without targeting tail generation directly. In all these works the decoder likelihood is either kept Gaussian or replaced by a fixed parametric family whose tail structure cannot adapt freely to data. The present paper isolates this as the decisive factor and shows that replacing the decoder with a non-parametric Phase-Type likelihood removes the structural constraint without modifying the encoder, prior, or training procedure.

\subsection{Phase-Type Distributions in Performance Evaluation}
 Within applied probability and performance modeling research fields it has been shown that PH distributions can be used to approximate heavy-tailed behavior. Authors in \cite{HoTe00} show that PH distributions can match both the body and moderate tail of distributions such as Pareto and lognormal by combining phases with progressively smaller exit rates. Feldman and Whitt~\cite{feldmann1998fitting} demonstrate that mixtures of exponentials, a subclass of acyclic PH, can be fitted to long-tailed network traffic data, providing a theoretical foundation for replacing exact heavy-tailed service times with PH approximations in M/PH/1 queues. More recently, we proposed hybrid Bernstein PH-hyperexponential models (BPH\_HE) ~\cite{ziani2025approximating} that separately handle the bulk and the tail of heavy-tailed distributions and validate them in queuing simulations. In this contribution, we extend this line of work to the deep generative setting: rather than fitting a PH distribution post-hoc to a known or empirically estimated distribution, we integrate PH distributions directly as the decoder likelihood of a VAE and train all parameters end-to-end via gradient descent through the exact matrix-exponential log-likelihood.
\section{Background}
\label{sec:background}
\subsection{Heavy-Tailed Distributions}
A distribution is called heavy-tailed if its complementary cumulative distribution function (CCDF), also known as the survival function, is defined as
$\bar{F}(x) = P(X > x)$, satisfies
\[
  \lim_{x \to \infty} e^{\lambda x} \bar{F}(x) = \infty \quad \forall \lambda > 0,
\]
meaning the tail decays more slowly than any exponential. A canonical subfamily is the class of \emph{regularly varying} distributions, whose tails decay in a polynomial way:
\[
  \bar{F}(x) \sim x^{-\alpha}, \quad \alpha > 0,
\]
By moving $\bar{F}(x) = P(X > x)$ above the alternative characterization, where $\alpha$ is the tail index. Smaller $\alpha$ implies a heavier tail and finite moments only up to order $\lfloor \alpha \rfloor - 1$. % this is the floor notation, apparently it is used to round down a number to closest lowest integer...

In this work, we use \emph{Pareto Type-I} distributions as a controlled testbed:
\[
  \bar{F}(x) = \left(\frac{x_m}{x}\right)^{\alpha}, \quad x \geq x_m > 0,
\]
with $x_m = 1$ throughout. The tail index $\alpha$ governs tail heaviness: $\alpha = 2$ yields an infinite-variance distribution, $\alpha = 3$ a finite variance but infinite third moment, $\alpha = 5$ a moderately heavy tail, and $\alpha = 30$ a distribution that is nearly Gaussian in behavior over typical sample ranges. Pareto samples are generated via the inverse CDF transform $X = x_m (1 - U)^{-1/\alpha}$, $U \sim \mathrm{Uniform}(0,1)$.

\subsection{Variational Autoencoders}
\label{sec:vae}
A Variational Autoencoder (VAE)~\cite{Kingma2013AutoEncodingVB} defines a latent-variable generative model $p_\theta(x, z) = p_\theta(x \mid z)\, p(z)$ with prior $p(z) = \mathcal{N}(0, I)$. An encoder network approximates the intractable posterior $p_\theta(z \mid x)$ with a variational distribution $q_\phi(z \mid x) = \mathcal{N}(z \mid \mu_\phi(x) , \mathrm{diag}(\sigma^2_\phi(x)))$. Model parameters are learned by maximizing the ELBO:
\[
  \mathcal{L}(\theta, \phi; x) = \mathbb{E}_{q_\phi(z|x)}\!\left[\log p_\theta(x \mid z)\right] - \mathrm{KL}\!\left[q_\phi(z \mid x) \;\|\; p(z)\right].
\]
For continuous data, the decoder likelihood $p_\theta(x \mid z)$ is almost universally chosen as a diagonal Gaussian:
\[
  p_\theta(x \mid z) = \mathcal{N}\!\left(x \mid \mu_\theta(z),\, \mathrm{diag}(\sigma^2_\theta(z))\right),
\]
where $\mu_\theta$ and $\sigma_\theta$ are neural networks \cite{DoerschX2016,LiangPan2024}. Under this choice, maximizing the ELBO reconstruction term is equivalent to minimizing a weighted mean squared error \cite{Diederik_2019,chen2017variationallossyautoencoder}, and the decoder mean $\mu_\theta(z)$ is the quantity typically used to represent generated samples.

\subsection{Phase-Type Distributions}

A Phase-Type (PH) distribution is the probability distribution of the time 
to absorption in a finite-state continuous-time Markov chain (CTMC) with $m$ 
transient states and one absorbing state~\cite{neuts1975}. An example is shown in Fig.~\ref{fig:ph} with three transient states. A PH distribution is compactly represented 
by a pair $(\boldsymbol{\alpha}, \mathbf{A})$, where $\boldsymbol{\alpha} \in 
\mathbb{R}^m$ is the initial probability vector over transient states and 
$\mathbf{A} \in \mathbb{R}^{m \times m}$ is the sub-generator matrix. The 
probability density function, CDF, and CCDF admit closed-form matrix-exponential 
expressions (which can be calculated efficiently by randomization \cite{Stewart_numerical_methods_markov_chains}):
\[
  f(x) = \boldsymbol{\alpha} e^{\mathbf{A}x}(-\mathbf{A}\mathbf{1}), \quad
  F(x) = 1 - \boldsymbol{\alpha} e^{\mathbf{A}x}\mathbf{1}, \quad 
  \bar{F}(x) = \boldsymbol{\alpha} e^{\mathbf{A}x}\mathbf{1}.
\]

\begin{wrapfigure}[12]{R}{0.38\textwidth}
\label{fig:ph}
  \vspace{-30pt}
  \begin{center}
    \includegraphics[width=0.36\textwidth]{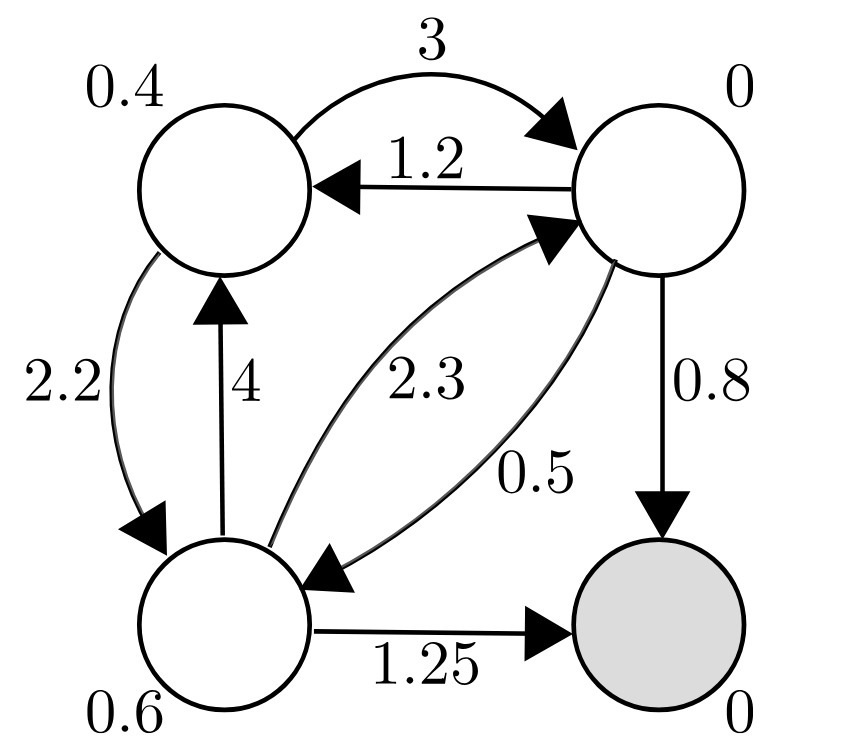}
  \end{center}
  \vspace{-5pt}
  \caption{A degree 3 PH distribution.\label{fig:ph}}
  \vspace{-10pt}
\end{wrapfigure}

PH distributions form a \emph{dense} family on $(0, \infty)$: any positive-valued distribution can be approximated arbitrarily well by a PH distribution. Although PH distributions are asymptotically light-tailed (their tails ultimately decay exponentially due to the finite Markov chain), they can closely approximate heavy-tailed behavior over any bounded, data-relevant range by using multiple phases with small exit rates~\cite{HoTe00}. This finite-range approximation is precisely what matters for data-driven generative modeling, where models are evaluated on observed data rather than asymptotic behavior.\\

\begin{figure}
    \centering
    \vspace{-30pt}
    \includegraphics[width=0.8\linewidth]{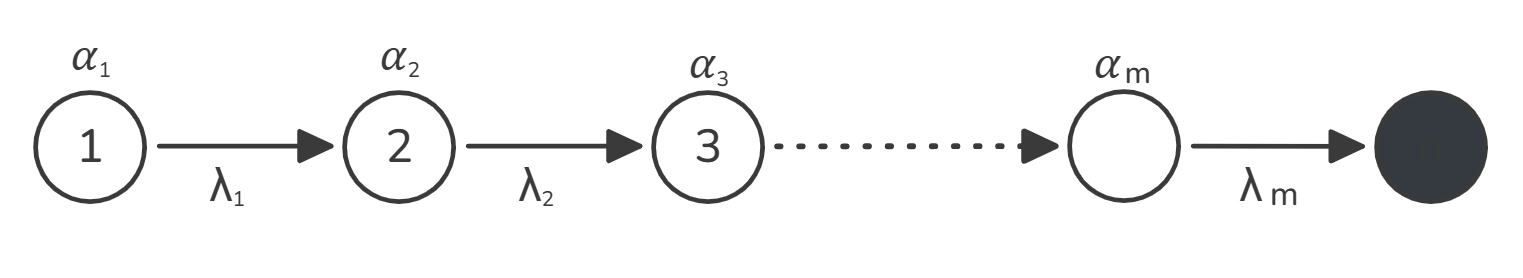}
    \caption{Series canonical form Phase-Type Distribution}
    \label{fig:scf_ph}
    \vspace{-30pt}
\end{figure}

For parameter efficiency and identifiability, we use acyclic PH distributions in the \emph{series canonical form} (shown in Fig.~\ref{fig:scf_ph}) where the sub-generator is upper bidiagonal with ordered rates $0 < \lambda_1 \leq \cdots \leq \lambda_m$:
\[
  \mathbf{A}_{ii} = -\lambda_i, \quad \mathbf{A}_{i,i+1} = \lambda_i, \quad \mathbf{A}_{ij} = 0 \text{ otherwise.}
\]
This representation requires only $2m-1$ parameters, which is equal to the true degrees of freedom of a PH distribution and eliminates non-identifiability~\cite{CUMANI1982583,asmussen2003applied}.

\section{Lipschitz Generative Models Fail on Heavy-Tailed Data}
\label{sec:theory}
We now provide a theoretical characterization of the tail-collapse phenomenon in the standard Gaussian VAE. The argument proceeds in two steps: first, we show that the Gaussian likelihood imposes an exponential tail on the decoder output, and second, we show that Lipschitz continuity of the decoder network prevents the latent mixing from compensating for this.

\subsection{The Gaussian Likelihood Constraint}
\label{sec:gaussiantheory}
The marginal distribution of generated samples is obtained by integrating 
the decoder likelihood over the latent prior:
\begin{equation}
p_\theta(x)
=
\int
\mathcal{N}\!\left(x \mid \mu_\theta(z), \sigma_\theta^2(z)\right)
\mathcal{N}(z \mid 0, I)\,dz.
\end{equation}

Thus, the model defines a continuous mixture of Gaussian distributions
indexed by the latent variable $z$. We now show that this marginal
distribution cannot produce heavy tails.

Consider the conditional distribution for a fixed $z$,
\[
X \mid z \sim \mathcal{N}\!\left(\mu_\theta(z), \sigma_\theta^2(z)\right).
\]
Define the conditional survival function
\[
\bar{F}(x \mid z) = P(X > x \mid z).
\]

Let
\[
t = \frac{x - \mu_\theta(z)}{\sigma_\theta(z)} .
\]
Then
\[
\bar{F}(x \mid z)
=
P\!\left(Z > t\right),
\quad
Z \sim \mathcal{N}(0,1).
\]

Using the classical Mills ratio inequality for the standard normal
distribution~\cite{grimmett2020probability},
\begin{equation}
\bar{\Phi}(t) \le \frac{\phi(t)}{t}, \qquad t>0,
\end{equation}
where $\phi(t)=\frac{1}{\sqrt{2\pi}}e^{-t^2/2}$ is the standard normal
density, we obtain
\begin{equation}
\bar{F}(x \mid z)
\le
\frac{\sigma_\theta(z)}{x-\mu_\theta(z)}
\,
\phi\!\left(
\frac{x-\mu_\theta(z)}{\sigma_\theta(z)}
\right).
\end{equation}

Substituting the expression of $\phi(\cdot)$ gives
\begin{equation}
\bar{F}(x \mid z)
\le
\frac{\sigma_\theta(z)}{\sqrt{2\pi}\,(x-\mu_\theta(z))}
\exp\!\left(
-\frac{(x-\mu_\theta(z))^2}{2\sigma_\theta^2(z)}
\right).
\end{equation}

Hence, the conditional survival probability decays exponentially in
$(x-\mu_\theta(z))^2$, which is much faster than any polynomial decay
$x^{-\alpha}$. Therefore, the conditional distribution is light-tailed
for every fixed $z$.

The marginal survival function is obtained by integrating over the
latent prior:
\begin{equation}
\bar{F}_\theta(x)
=
\int \bar{F}(x \mid z)\,p(z)\,dz.
\end{equation}

Since each conditional tail decays exponentially and the latent prior $p(z) = \mathcal{N}(0, I)$ has exponentially decaying tails, the marginal
distribution $\bar{F}_\theta(x)$ also decays faster than any polynomial
in $x$. The resulting distribution is therefore light-tailed.

The argument relies on the fact that the decoder outputs
$\mu_\theta(z)$ and $\sigma_\theta(z)$ remain finite for almost all
$z \sim p(z)$. In practice, this is ensured by the Lipschitz structure
of neural network decoders, which prevents arbitrarily large growth of
$\mu_\theta(z)$ with respect to $z$.

\subsection{The Lipschitz Amplification Bound}

The conclusion of Section \ref{sec:gaussiantheory} does not yet by itself rule out heavy tails in the
marginal distribution, because the Gaussian components are mixed over the latent
variable $z$. In principle, one might hope that extreme latent values could be
mapped by the decoder to sufficiently large means $\mu_\theta(z)$ so that the
mixture recovers a heavy tail. We now show that this cannot happen when the
decoder is Lipschitz and the latent prior is Gaussian.

Let $\mu_\theta : \mathbb{R}^{d} \to \mathbb{R}^{D}$ denote the decoder mean
network where $d$ is the dimension of the latent variable $z$ and $D$ is the
dimension of the observed data $x$. Since it is a feed-forward neural network with bounded weights and
Lipschitz activations, it is itself Lipschitz-continuous: there exists a constant
$L>0$ such that
\begin{equation}
\|\mu_\theta(z)-\mu_\theta(z')\| \le L \|z-z'\|,
\qquad \forall z,z' \in \mathbb{R}^{d}.
\end{equation}

Taking $z'=0$ yields
\begin{equation}
\|\mu_\theta(z)\|
\le
\|\mu_\theta(0)\| + L\|z\|.
\end{equation}
Hence, the decoder mean can grow at most linearly with the latent variable.
In particular, for any scalar output coordinate $\mu_{\theta,j}(z)$,
\begin{equation}
|\mu_{\theta,j}(z)|
\le
\|\mu_\theta(z)\|
\le
b + L\|z\|,
\qquad
b := \|\mu_\theta(0)\|.
\end{equation}

Now consider the event that the decoder mean reaches a large level $x>0$.
If $\mu_{\theta,j}(z) > x$, then necessarily
\begin{equation}
b + L\|z\| > x,
\end{equation}
which implies
\begin{equation}
\|z\| > \frac{x-b}{L}.
\end{equation}
Therefore,
\begin{equation}
P\!\left(\mu_{\theta,j}(z)>x\right)
\le
P\!\left(\|z\|>\frac{x-b}{L}\right).
\end{equation}

Since the latent prior is Gaussian, $z \sim \mathcal{N}(0,I)$, it has
light (sub-Gaussian) tails, so large values of $\|z\|$ are exponentially
unlikely. Using the Lipschitz bound $\mu_{\theta,j}(z) \le L\|z\| + b$,
large values of $\mu_{\theta,j}(z)$ require proportionally large latent
norms. In particular, reaching an output level $x$ requires
$\|z\| \gtrsim (x-b)/L$, which occurs with exponentially small
probability under the Gaussian prior. Hence the random variable
$\mu_{\theta,j}(z)$ induced by the latent prior is itself light-tailed.

This resolves the remaining gap in Section~\ref{sec:gaussiantheory}.
The conditional Gaussian tail bound depends on the shifted quantity
$x-\mu_\theta(z)$. For the marginal distribution to exhibit a heavy
tail, the decoder would need to produce values of $\mu_\theta(z)$ that
track $x$ frequently enough to compensate for the exponential decay of
the Gaussian likelihood. The Lipschitz constraint makes this impossible:
The mean can only grow linearly with $\|z\|$, while the Gaussian prior
assigns exponentially small probability to the latent values required
to push $\mu_\theta(z)$ toward $x$.
Consequently, latent mixing cannot compensate for the Gaussian likelihood
constraint. The decoder may translate or rescale Gaussian components, but it
cannot amplify the light-tailed Gaussian prior into a distribution whose
tail is decaying in a polynomial way. The marginal output distribution, therefore, remains light-tailed.

\subsection{Phase-Type Decoder}
A Phase-Type decoder replaces the Gaussian conditional $p_\theta(x \mid z)$
with a PH distribution $\mathrm{PH}(\boldsymbol{\alpha}(z), \mathbf{A}(z))$
whose parameters are outputs of neural networks. Crucially, the tail behavior
of a PH distribution is governed by the eigenvalue of $\mathbf{A}(z)$ with the
largest real part (i.e., the one closest to zero). Denoting this eigenvalue by
$-\lambda_1(z)$ with $\lambda_1(z)>0$, the survival function satisfies
$\bar F(x)~\sim~C e^{-\lambda_1(z)x}$ as $x\to\infty$.

Although PH distributions are still light-tailed in the strict asymptotic
sense, small values of $\lambda_1(z)$ produce very slow exponential decay,
which yields an increasingly heavy effective tail over a large finite
ranges \cite{HoTe00}. Since the decoder network can output arbitrarily small
(but positive) values of $\lambda_1(z)$ for suitable latent inputs, the PHThe 
decoder is not subject to the Gaussian tail constraint.

While the Lipschitz constraint still applies to the parameter outputs
$\boldsymbol{\alpha}(z)$ and $\mathbf{A}(z)$, the mapping from these parameters
The resulting tail behavior is highly nonlinear: small changes in
$\lambda_1(z)$ near zero induces large changes in the effective tail decay.
This asymmetry enables PH decoders to approximate heavy-tailed behavior that
Gaussian decoders cannot capture.
\section{Experimental Setup}
\label{sec:setup}

\subsection{Data Generation}
We generate synthetic datasets from independent multivariate Pareto Type-I distributions:
\[
  X \sim \mathrm{Pareto}(\alpha, x_m), \quad x_m = 1,
\]
with tail index $\alpha \in \{2, 3, 5, 30\}$ and dimensionality $d \in \{1, 5, 10\}$, yielding a $4 \times 3$ experimental grid of 12 configurations. Each dimension is drawn independently. Dataset sizes are: training set $N_\mathrm{train} = 20{,}000$, and generated set $N_\mathrm{gen} = 20{,}000$.

The tail index grid is designed to span qualitatively distinct regimes: $\alpha = 2$ corresponds to an infinite-variance distribution; $\alpha = 3$ to finite variance but infinite skewness; $\alpha = 5$ to a moderately heavy tail with finite low-order moments; and $\alpha = 30$ to a near-Gaussian distribution that serves as a boundary condition and sanity check.
\subsection{Models}
\paragraph{Gaussian VAE (baseline).}
The encoder and decoder are both multilayer perceptrons (MLPs) with 2 hidden layers of size 64, ReLU activations, and latent dimension $d_z = 8$. The decoder outputs the mean and log-variance of a diagonal Gaussian:
\[
  p_\theta(x \mid z) = \mathcal{N}(\mu_\theta(z), \mathrm{diag}(\sigma^2_\theta(z))).
\]

\paragraph{PH-VAE.}
Identical encoder architecture and latent dimension. The decoder outputs the parameters of a Phase-Type distribution in series canonical form with $m = 10$ phases per dimension. Specifically, the decoder outputs:
\begin{itemize}
  \item An initial probability vector $\boldsymbol{\alpha}_j(z) \in \mathbb{R}^m$ per dimension $j$, enforced via softmax.
  \item A vector of positive, ordered rates $\boldsymbol{\lambda}_j(z)$, enforced via softplus followed by cumulative sum, guaranteeing $0 < \lambda_{j,1} \leq \cdots \leq \lambda_{j,m}$.
\end{itemize}
The conditional likelihood for dimension $j$ is:
\[
  p_\theta(x_j \mid z) = \boldsymbol{\alpha}_j(z)\, \exp(\mathbf{A}_j(z)\, x_j)\, (-\mathbf{A}_j(z)\mathbf{1}),
\]
evaluated efficiently via uniformization~\cite{Stewart_numerical_methods_markov_chains}. The full decoder factorizes across dimensions conditionally on $z$, inducing cross-dimensional dependence through the shared latent representation.

\paragraph{Training.}
Both models are trained with the Adam optimizer, learning rate $10^{-3}$, batch size 128, for 100 epochs. The ELBO training objective for the PH-VAE replaces the Gaussian log-likelihood with the exact PH log-likelihood. All other training settings are identical.

\subsection{Evaluation Metrics}
We evaluate both global distribution fit and, more importantly, tail fidelity.
\paragraph{Tail-KS.}
To isolate tail behavior, we define a conditional Kolmogorov-Smirnov (KS) distance restricted to the upper tail. Let $u = Q_{0.99}(X_\mathrm{test})$ be the empirical 99th percentile of the test set. We compute:
\[
  \mathrm{KS}_\mathrm{tail} = \sup_{x \geq u} \left| F_\mathrm{gen}(x \mid x > u) - F_\mathrm{test}(x \mid x > u) \right|,
\]
where conditional CDFs are normalized to integrate to 1 above $u$. A value of $\mathrm{KS}_\mathrm{tail} = 1.0$ indicates complete tail collapse: the generated distribution places no mass above the test 99th percentile.

\paragraph{Extreme Quantile Error.}
For extreme quantile levels $q \in \{0.99, 0.995\}$, we compute the relative error:
\[
  Q_\mathrm{err}(q) = \frac{|Q_\mathrm{gen}(q) - Q_\mathrm{test}(q)|}{Q_\mathrm{test}(q)}.
\]

\section{Results}
\label{sec:results}

\subsection{Tail Fidelity: Survival Function Analysis}
\label{sec:ccdf}

\begin{figure}[t]
    \centering
    \begin{minipage}[t]{0.48\textwidth}
        \centering
        \includegraphics[width=\textwidth]{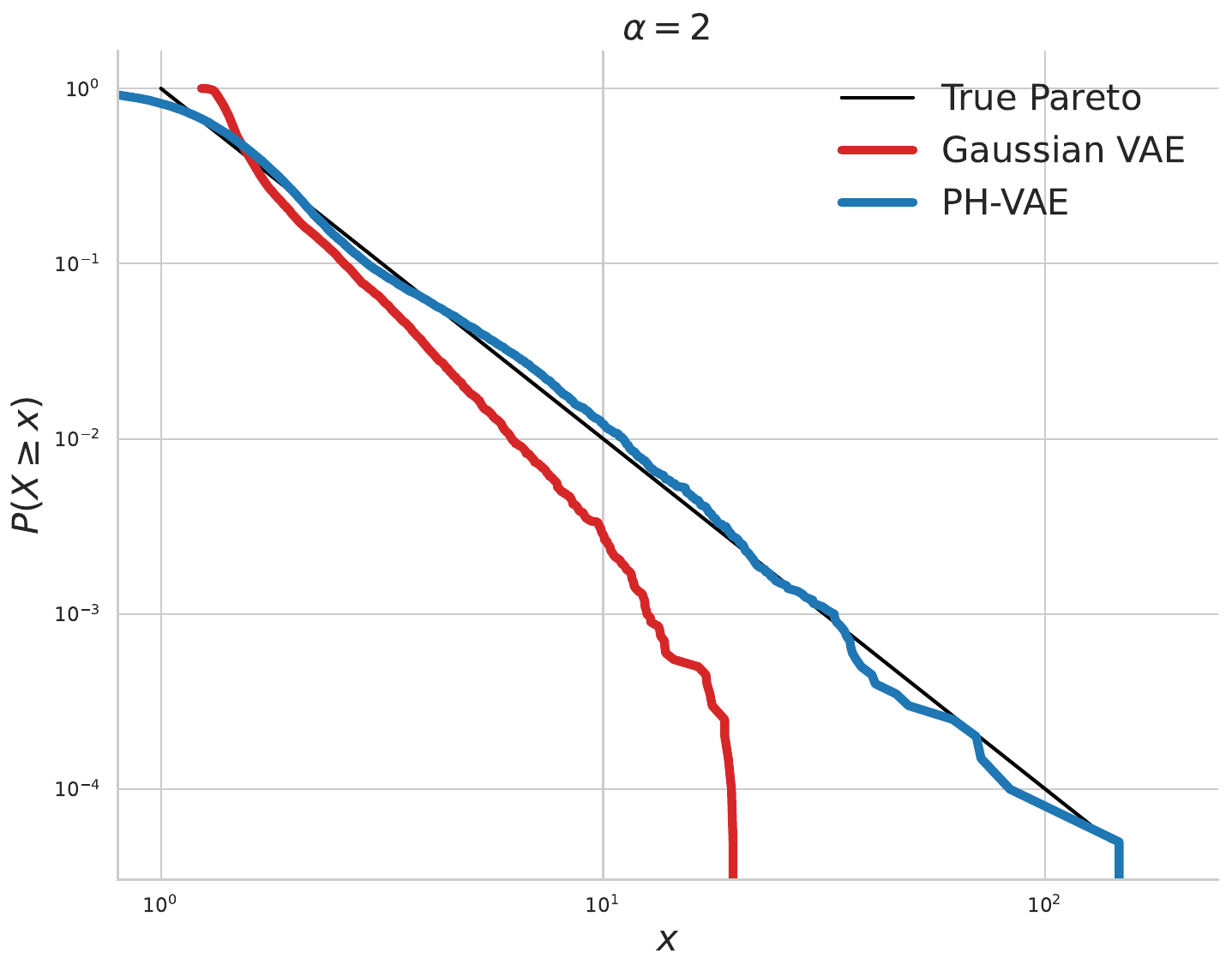}
    \end{minipage}
    \hfill
    \begin{minipage}[t]{0.48\textwidth}
        \centering
        \includegraphics[width=\textwidth]{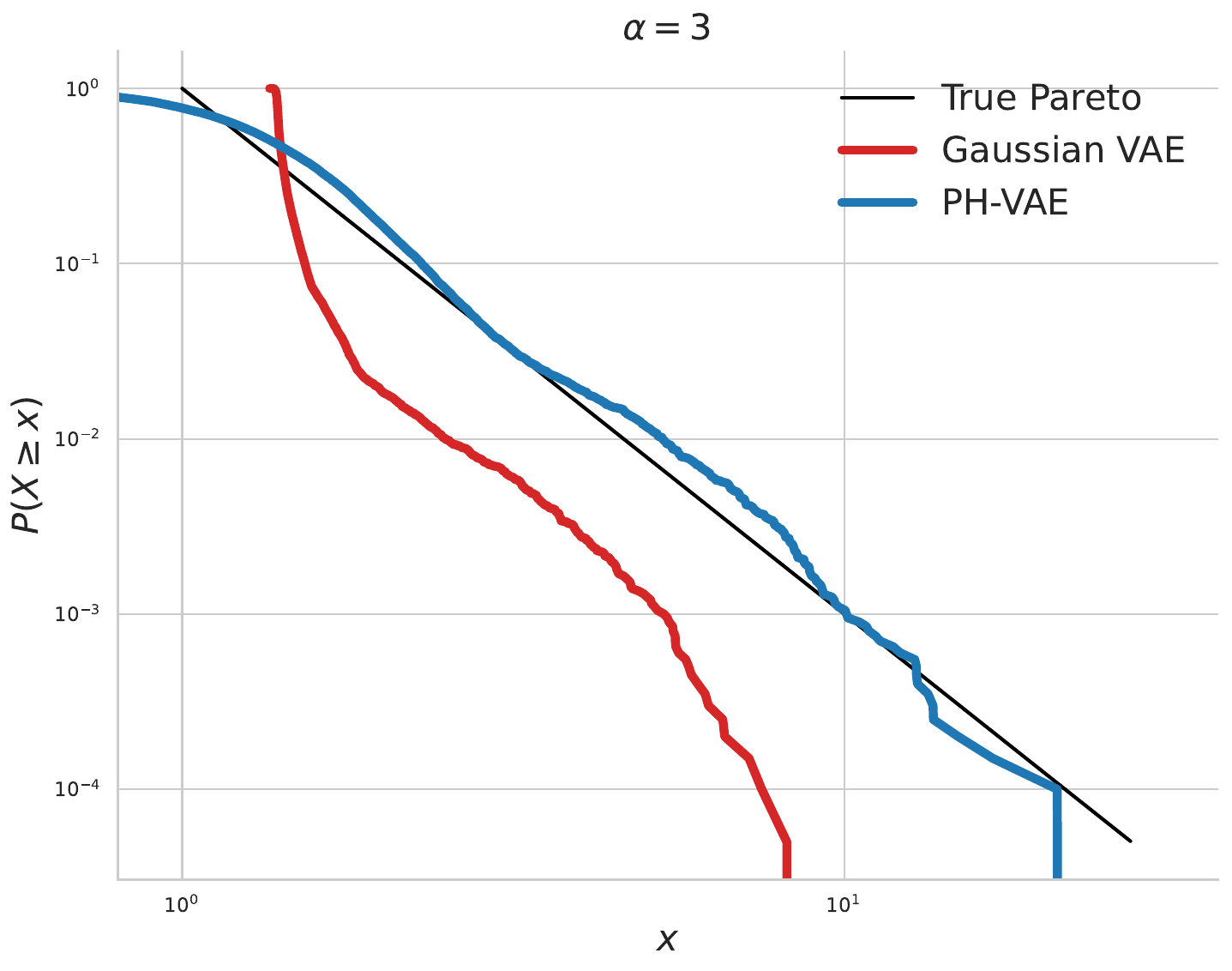}
    \end{minipage}
    \caption{Log-log CCDF of true Pareto data, Gaussian VAE generations,
    and PH-VAE generations at dimension $d=1$. \textit{Left:} $\alpha=2$. \textit{Right:}  $\alpha=3$.}
    \label{fig:ccdf}
\end{figure}

Figure \ref{fig:ccdf} shows the log-log
CCDF for both models against the true Pareto distribution at $\alpha=2$
and $\alpha=3$, dimension $d=1$. Any generated curve that drops below the true CCDF is underestimating the probability of extreme values.

At $\alpha=2$, both models track the true distribution in the bulk, but
the Gaussian VAE diverges visibly in the upper tail, its curve bends
downward around $x \approx 5$ and runs out of generated mass near $x=20$.
The PH-VAE stays close to the true CCDF throughout, including in the far
tail beyond $x=100$.

At $\alpha=3$ the contrast is starker. The Gaussian VAE drops sharply
away from the true line and terminates near $x=5$, generating nothing
in the region where the Pareto distribution still assigns probability
on the order of $10^{-3}$ to $10^{-4}$. This is the complete tail
collapse reported as $\mathrm{KS}_\mathrm{tail}=1.0$ in Table~\ref{tab:results}.
The PH-VAE continues to follow the true slope well into the tail, only
diverging slightly beyond $x=10$ where sample noise becomes significant.

\subsection{Quantitative Evaluation}
\label{sec:quant}
\begin{figure}[t]
    \centering
    \begin{tabular}{cc}
        \includegraphics[width=0.454\textwidth]{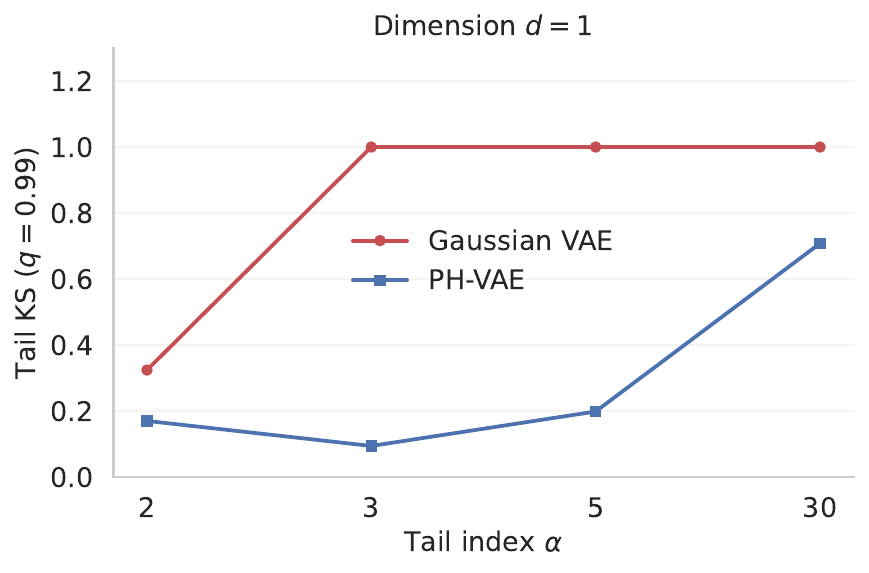} & 
        \includegraphics[width=0.454\textwidth]{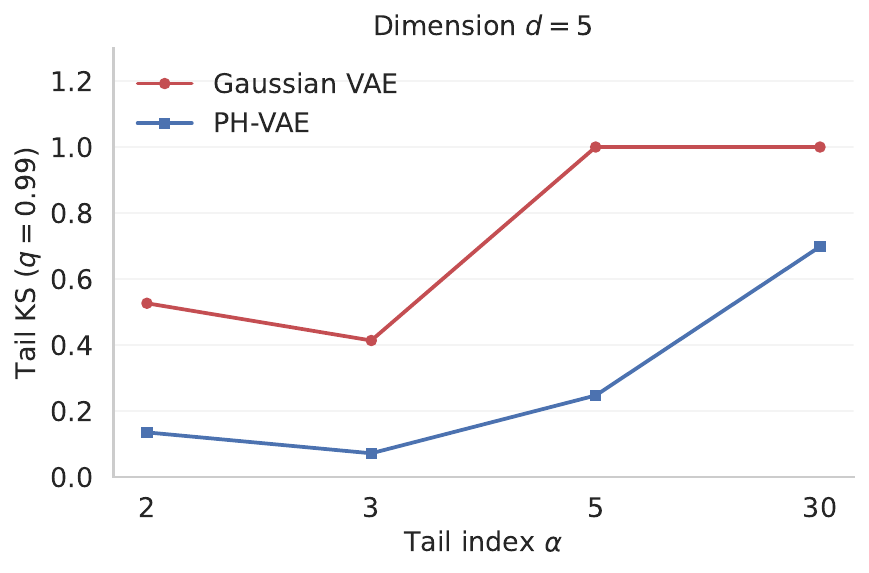} \\
        \includegraphics[width=0.45\textwidth]{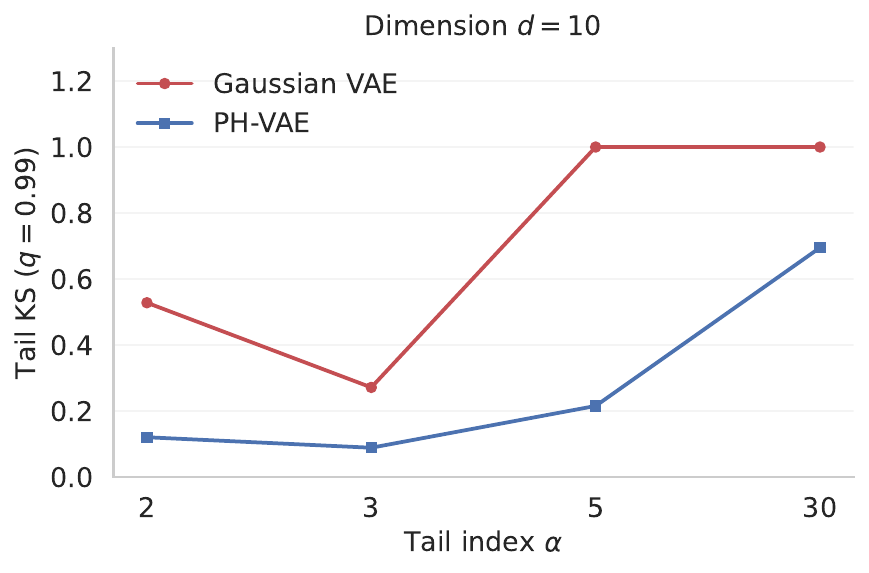} \\
    \end{tabular}
    \caption{Tail KS distance at the 99th percentile for Gaussian VAE and PH-VAE across Pareto tail indices.}
    \label{fig:ks_alpa}
\end{figure}
Table~\ref{tab:results} reports KS, $\mathrm{KS}_\mathrm{tail}$, and extreme quantile errors across the full $\alpha \times d$ grid. Figure~\ref{fig:ks_alpa} shows $\mathrm{KS}_\mathrm{tail}$ as a function of $\alpha$ for each dimension, making the progression of tail collapse across the tail index grid visually explicit.

\begin{table}
\centering
\caption{Evaluation metrics across the full $\alpha \times d$ experimental grid. $\mathrm{KS}_\mathrm{tail} = 1.0$ indicates complete tail collapse (no generated mass above the test 99th percentile). Best result per configuration shown in \textbf{bold}.}
\label{tab:results}
\setlength{\tabcolsep}{4.5pt}
\begin{tabular}{ccl cccc}
\toprule
$\alpha$ & $d$ & Model & KS $\downarrow$ & KS$_\mathrm{tail}$ $\downarrow$ & $Q_{0.99}$ err $\downarrow$ & $Q_{0.995}$ err $\downarrow$ \\
\midrule
\multirow{6}{*}{2} 
 & \multirow{2}{*}{1}  & Gaussian VAE & 0.267 & 0.324 & 0.184 & 0.297 \\
 &                     & PH-VAE       & \textbf{0.189} & \textbf{0.170} & \textbf{0.045} & \textbf{0.041} \\
\cmidrule{2-7}
 & \multirow{2}{*}{5}  & Gaussian VAE & 0.251 & 0.526 & 0.373 & 0.469 \\
 &                     & PH-VAE       & \textbf{0.177} & \textbf{0.135} & \textbf{0.067} & \textbf{0.105} \\
\cmidrule{2-7}
 & \multirow{2}{*}{10} & Gaussian VAE & 0.277 & 0.528 & 0.164 & 0.316 \\
 &                     & PH-VAE       & \textbf{0.162} & \textbf{0.120} & \textbf{0.074} & \textbf{0.114} \\
\midrule
\multirow{6}{*}{3}
 & \multirow{2}{*}{1}  & Gaussian VAE & 0.560 & 1.000 & 0.509 & 0.483 \\
 &                     & PH-VAE       & \textbf{0.231} & \textbf{0.094} & \textbf{0.101} & \textbf{0.092} \\
\cmidrule{2-7}
 & \multirow{2}{*}{5}  & Gaussian VAE & 0.549 & 0.414 & 0.328 & 0.364 \\
 &                     & PH-VAE       & \textbf{0.210} & \textbf{0.072} & \textbf{0.150} & \textbf{0.131} \\
\cmidrule{2-7}
 & \multirow{2}{*}{10} & Gaussian VAE & 0.503 & 0.271 & 0.327 & 0.332 \\
 &                     & PH-VAE       & \textbf{0.233} & \textbf{0.089} & \textbf{0.070} & \textbf{0.074} \\
\midrule
\multirow{6}{*}{5}
 & \multirow{2}{*}{1}  & Gaussian VAE & 0.643 & 1.000 & 0.498 & 0.569 \\
 &                     & PH-VAE       & \textbf{0.335} & \textbf{0.198} & \textbf{0.050} & \textbf{0.011} \\
\cmidrule{2-7}
 & \multirow{2}{*}{5}  & Gaussian VAE & 0.648 & 1.000 & 0.500 & 0.568 \\
 &                     & PH-VAE       & \textbf{0.342} & \textbf{0.247} & \textbf{0.032} & \textbf{0.034} \\
\cmidrule{2-7}
 & \multirow{2}{*}{10} & Gaussian VAE & 0.637 & 1.000 & 0.498 & 0.563 \\
 &                     & PH-VAE       & \textbf{0.325} & \textbf{0.215} & \textbf{0.055} & \textbf{0.027} \\
\midrule
\multirow{6}{*}{30}
 & \multirow{2}{*}{1}  & Gaussian VAE & {0.514} & 1.000 & \textbf{0.104} & \textbf{0.126} \\
 &                     & PH-VAE       & \textbf{0.496} & \textbf{0.708} & 0.998 & 1.100 \\
\cmidrule{2-7}
 & \multirow{2}{*}{5}  & Gaussian VAE & 0.583 & 1.000 & \textbf{0.109} & \textbf{0.131} \\
 &                     & PH-VAE       & \textbf{0.521} & \textbf{0.698} & 0.897 & 0.990 \\
\cmidrule{2-7}
 & \multirow{2}{*}{10} & Gaussian VAE & 0.575 & 1.000 & \textbf{0.108} & \textbf{0.129} \\
 &                     & PH-VAE       & \textbf{0.510} & \textbf{0.695} & 0.861 & 0.957 \\
\bottomrule
\end{tabular}
\end{table}

\paragraph{Heavy-tail regime ($\alpha = 2, 3$).}
For the heaviest tails, the PH decoder provides large and consistent improvements across all metrics and dimensions. At $\alpha = 2$, the Gaussian VAE achieves $\mathrm{KS}_\mathrm{tail}$ between 0.32 and 0.53, while the
PH-VAE reduces this to between 0.12 and 0.17, a factor of 2--4 improvement. Extreme quantile errors are reduced more dramatically: $Q_{0.99}$ error drops from 0.18--0.37 to 0.04--0.07, and $Q_{0.995}$ error from 0.30--0.47 to 0.04--0.11. At $\alpha = 3$, the Gaussian VAE reaches $\mathrm{KS}_\mathrm{tail} = 1.0$ at $d=1$, which is a complete tail collapse, while the PH-VAE achieves $\mathrm{KS}_\mathrm{tail} = 0.094$, a factor of more than 10 improvement. The $Q_{0.99}$ errors for the Gaussian VAE at $\alpha=3$(0.33--0.51) indicate systematic underestimation of extreme quantiles by 33--51\%, a dangerous property for any risk-sensitive application such as queue dimensioning or traffic engineering.

\section{Discussion}
\label{sec:discussion}
The experimental results confirm the theoretical prediction of Section~\ref{sec:theory}. The Gaussian VAE exhibits complete tail collapse ($\mathrm{KS}_\mathrm{tail}=1.0$) in 9 out of 12 configurations, all with $\alpha \geq 3$. This failure is structural: due to its Gaussian likelihood, the model cannot generate values beyond the empirical 99th percentile.

This structural argument has practical implications for the performance evaluation community, where heavy-tailed service and inter-arrival times are common. A generative model trained on such data and used to synthesize traffic or workloads will systematically underestimate extreme values, potentially leading to under-provisioned systems.

The performance evaluation community has long used PH distributions as a tractable tool for approximating heavy-tailed distributions in queuing models~\cite{ziani2025approximating}. Our results extend this intuition to the setting of deep generative models: PH distributions are equally effective as decoder likelihoods learned end-to-end from data, not just as fitted approximations to known distributions. Our hybrid BPH-HE models of~\cite{ziani2025approximating} demonstrate that PH distributions can approximate both the body and the tail of heavy-tailed distributions with high accuracy. Our contribution is to show that these approximation properties are preserved and remain beneficial when PH distributions are integrated into a deep generative model trained end-to-end via gradient descent.

\subsection{Limitations and Future Work}
The current study has some limitations. Experiments use synthetic Pareto data only, which provides clean control over tail heaviness but leaves open how the PH decoder performs on real heavy-tailed data such as network inter-arrival times, queue service times, or insurance claim sizes. Future work includes evaluation on real queuing and network traffic datasets and extending the PH decoder to data supported on $\mathbb{R}$ rather than $\mathbb{R}_+$.
\section{Conclusion}
\label{sec:conclusion}

Standard VAEs fail on heavy-tailed data for a structural reason: the
combination of a Gaussian decoder and a Lipschitz neural network places a hard ceiling on how heavy the generated tail can be. No amount of additional capacity or longer training can overcome this, the model is not expressive enough in the tail by construction. We show that replacing the decoder with a Phase-Type distribution removes this ceiling. Because tail heaviness in a PH distribution is controlled by the smallest exit rate $\lambda_1(z)$, which the decoder network can push arbitrarily close to zero, the PH decoder escapes the structural constraint that defeats the Gaussian. The conducted experiments confirm this across a grid of Pareto tail indices and dimensionalities: the Gaussian VAE collapses completely in 9 out of 12 configurations, while the PH-VAE reduces tail KS distance and extreme quantile errors by up to a factor of 10. 

For the performance evaluation community, the implication is direct:
PH distributions, already the standard tool for approximating
Heavy-tailed service times in queuing models are equally effective
as learned decoder likelihoods in deep generative models.

\bibliographystyle{elsarticle-num}
\bibliography{main}
\end{document}